\documentclass{edm_article}
\usepackage{url}
\usepackage[T1]{fontenc}
\usepackage{graphicx}
\usepackage{amsmath}
\usepackage{booktabs}
\usepackage{xcolor}
\usepackage{amssymb}
\usepackage{listings}
\newcommand{\gabriel}[1]{\textcolor{red}{[Gabriel: #1]}}
\newcommand{\lolo}[1]{\textcolor{orange}{[Lorenzo: #1]}}
\newcommand{\richb}[1]{\textcolor{green}{[RichB: #1]}}
\newcommand{\dbm}[1]{\textcolor{teal}{[Debshila: #1]}}
\newcommand{\arxiv}[1]{arXiv:#1}

\renewcommand{\gabriel}[1]{}
\renewcommand{\lolo}[1]{}
\renewcommand{\richb}[1]{}
\renewcommand{\dbm}[1]{}
\renewcommand{\arxiv}[1]{}

\usepackage[mode=buildnew]{standalone}
\usepackage{tikz}

\begin{document}


\title{Stable and Privacy-Preserving Synthetic Educational Data with Empirical Marginals: A Copula-Based Approach}

\numberofauthors{4}
\author{
\alignauthor
Gabriel Diaz Ramos\\
       \affaddr{Rice University}\\
       \email{gd27@rice.edu}
\alignauthor
Lorenzo Luzi\\
       \affaddr{Rice University}\\
       \email{ll65@rice.edu}
\alignauthor
Debshila Basu Mallick\\
       \affaddr{OpenStax}\\
       \email{db19@rice.edu}
\and
\alignauthor
Richard G. Baraniuk\\
       \affaddr{Rice University, OpenStax}\\
       \email{richb@rice.edu}
}

\maketitle

\begin{abstract}
To advance Educational Data Mining (EDM) within strict privacy-protecting regulatory frameworks, researchers must develop methods that enable data-driven analysis while protecting sensitive student information. 
Synthetic data generation is one such approach, enabling the release of statistically generated samples instead of real student records; however, existing deep learning and parametric generators often distort marginal distributions and degrade under iterative regeneration, leading to distribution drift and progressive loss of distributional support that compromise reliability. 
In response, we introduce the Non-Parametric Gaussian Copula (NPGC), a plug-and-play synthesis method that replaces deep learning and parametric optimization with empirical statistical anchoring to preserve the observed marginal distributions while modeling dependencies through a copula framework. 
NPGC integrates Differential Privacy (DP) at both the marginal and correlation levels, supports heterogeneous variable types, and treats missing data as an explicit state to retain informative absence patterns. 
We evaluate NPGC against deep learning and parametric baselines on five benchmark datasets and demonstrate that it remains stable across multiple regeneration cycles and achieves competitive downstream performance at substantially lower computational cost. 
We further validate NPGC through deployment in a real-world online learning platform, demonstrating its practicality for privacy-preserving research.

\end{abstract}


\section{Introduction}
\label{intro}

Educational Data Mining (EDM) and Learning Analytics (LA) rely on student-level records such as interaction logs, assessment outcomes, and contextual variables, governed by strict privacy and data-protection regulations. Even after removing direct identifiers, education datasets may remain disclosive because fine-grained behavioral traces and contextual variables enable re-identification or harm \cite{Klose_undated-dp}. This constraint is evident in the field’s open-science record: A reproducibility audit of LAK proceedings (11th–12th conferences) found that 5\% of papers shared raw datasets and 2\% reported data available “on request” \cite{Haim2023-qm}. A parallel audit of EDM proceedings (14th–15th conferences) found that 15\% of papers used or shared open datasets, 5\% reported availability “on request,” and most provided no accessible pathway to datasets \cite{2023.EDM-long-papers.10}. Together, these findings indicate a structural constraint on cumulative progress in EDM and LA: The data that power state-of-the-art models are often the least shareable, limiting reproducibility, benchmarking, and cross-institution comparison.

\begin{figure}[!t]
    \centering
    \Description{}
    \includegraphics[width=0.95\linewidth]{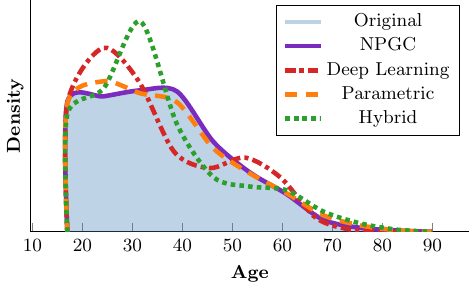}
    \caption{Marginal density of the \textit{Age} variable in the Adult~\cite{Adult} dataset. The shaded region denotes the empirical density of the real data. Curves correspond to synthetic samples generated by a deep learning model, a Parametric Gaussian Copula (PGC), a Copula-GAN hybrid, and our proposed Non-Parametric Gaussian Copula (NPGC). Existing methods exhibit visible deviations from the empirical marginal, whereas NPGC aligns perfectly with the real distribution.}
    \label{fig:problem_image}
\end{figure}

The impact of this constraint is increasing with the rapid uptake of generative AI (genAI), particularly large language models (LLMs) \cite{Vaswani2017}, which increases demand for high-quality student datasets, including open-ended responses, code, and dialogue. A recent systematic review of empirical LLM-in-education studies (Nov 2022–Mar 2025) documents rapid deployment while highlighting recurring privacy and governance challenges \cite{Shi2026LLMEducationReview}. Many LLM-based learning supports (e.g., tutor-style feedback and hint generation) are validated on real student work and interaction traces that are difficult to release safely at scale due to re-identification risk \cite{Phung2024-lz}. The model-development pipeline introduces additional exposure: Fine-tuning and adaptation rely on sensitive datasets and admit privacy attacks (e.g., membership inference and data extraction), motivating stronger privacy-preserving practices \cite{Du2025-jc}. This tension between data-hungry models and privacy-constrained records makes synthetic da\-ta generation compelling for EDM and LA, enabling method development and sharing while reducing reliance on protected student records and motivating the methodology proposed in this paper.

Synthetic data generation for tabular data (SDG-T) provides a practical approach for addressing this constraint. By combining statistical modeling with Differential Privacy (DP) \cite{DP} and established privacy evaluation protocols~\cite{privacy2012}, SDG-T constructs artificial datasets that preserve statistical structure without exposing individual records. SDG-T enables privacy-compliant analysis and model development in educational settings. DP is a formal framework that limits how much any single individual’s data can influence released statistics, providing provable guarantees of record-level privacy.

Despite its promise, SDG-T faces a central technical challenge: preserving the empirical marginal distributions of heterogeneous variables. We refer to deviations between real and synthetic feature distributions as \emph{marginal mismatch}. Educational attributes—such as test scores, attendance rates, or financial aid counts—rarely follow standard parametric families and often exhibit skewness, multi-mo\-dal\-ity, and discrete mass points. 

Deep learning generators, including TVAE~\cite{Synthcity} and CTGAN~\cite{CTGAN_TVAE}, optimize joint objectives that prioritize inter-column dependencies, often causing systematic errors like mode collapse~\cite{Kossale2022}. These models can distort individual marginals even when aggregate utility metrics appear competitive. Parametric approaches, such as the Gaussian Copula~\cite{SklarCopulas}, assume predefined marginal distributions that may not reflect the real data shape. Hybrid methods, including Copula-GAN~\cite{SDV}, apply Gaussianization before adversarial training, but still rely on fitted marginal models and do not guarantee exact empirical recovery.

As shown in Fig.~\ref{fig:problem_image}, existing generators often distort individual feature marginals relative to the empirical data. Deep learning models lack explicit marginal constraints, whereas parametric approaches impose distributional forms that may not reflect the true distribution of the variables. In contrast, NPGC preserves empirical marginals through non-parametric anchoring. Marginal drift can bias downstream analysis; this example reflects behavior observed consistently across datasets in our experiments.

Marginal distortion is further amplified under iterative regeneration. Because synthetic data can be produced at scale~\cite{pollution}, subsequent models may be trained on previously generated samples rather than on the original dataset~\cite{Shumailov2024}. We refer to this recursive setting as a Synthetic Feedback Loop (SFL). To quantify robustness under SFL, we define regeneration stability as the preservation of statistical structure across iterations. Under repeated regeneration, SFL can produce progressive variance shrinkage and loss of distributional support, consistent with \emph{model collapse}~\cite{Alemohammad2024}. 

\begin{figure}[t]
     \centering
     \Description{}
     \includegraphics[width=0.95\linewidth]{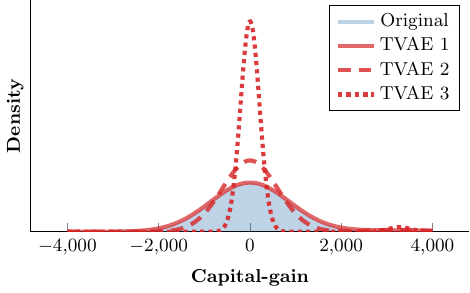}
    \caption{Marginal density of the \textit{Capital-gain} variable in the Adult~\cite{Adult} dataset across three sequential regeneration iterations of a Tabular Variational Autoencoder (TVAE). The shaded region denotes the empirical density of the original data, and red curves (iterations 1–3) represent synthetic samples generated under repeated synthetic feedback. The progressive concentration of probability mass indicates variance collapse and loss of distributional support across iterations.}
     \label{fig:mode_collapse_intro}
\end{figure}

We examine this SFL phenomenon in tabular synthesis. As shown in Fig.~\ref{fig:mode_collapse_intro}, unconstrained regeneration leads to systematic deviation from the original distribution across iterations. Such degradation is particularly consequential in educational datasets, where sub-representation of low-frequency groups can bias subgroup analysis and downstream prediction.

To address both marginal mismatch and regeneration instability, we introduce the Non-Parametric Gaussian Copula (NPGC). NPGC enforces exact empirical marginal preservation through non-parametric anchoring while modeling inter-variable dependencies in a Gaussian copula framework. The method supports continuous, integer, and categorical variables, and treats missing values as informative signals when estimating marginals and correlations. 

NPGC integrates DP by injecting calibrated noise into both the marginal distributions and the correlation structure, providing formal privacy guarantees. By constraining synthetic generation to respect empirical marginals, NPGC mitigates variance degradation under SFL while maintaining competitive downstream utility and substantially lower computational cost than deep learning–based baselines. While NPGC improves marginal fidelity and regeneration stability, it models dependencies through a Gaussian copula on rank-trans\-formed data. NPGC captures a broad class of relationships, but some complex nonlinear interactions may not be fully preserved.

We evaluate NPGC with respect to fidelity, privacy, regeneration stability, and downstream predictive utility, each formally defined in Section~\ref{experiments_explained}. Fidelity measures statistical agreement between real and synthetic data, privacy quantifies resistance to disclosure and inference attacks, and regeneration stability evaluates robustness under iterative synthesis. 

Experiments are conducted on five datasets from the UCI Machine Learning Repository~\cite{Uci}, a widely used benchmark collection of heterogeneous tabular datasets, including two education-focused datasets (\textit{Student Dropout}~\cite{Dropout}, and \textit{Student Performance}~\cite{Performance}). Figures~\ref{fig:problem_image} and~\ref{fig:mode_collapse_intro} use the \textit{Adult}~\cite{Adult} dataset due to its larger sample size, which enables clear visualization of marginal behavior and regeneration effects.

Across benchmarks, NPGC achieves higher fidelity and privacy scores than baseline methods while maintaining competitive utility and substantially lower computational cost. By enforcing empirical marginals, NPGC stabilizes regeneration under SFL and mitigates variance degradation.

We further evaluate NPGC in a real-world deployment using interaction logs from a large-scale online learning platform.\footnote{Specific details regarding the online platform and workshop venue have been anonymized for the peer-review process.} The dataset exhibits severe categorical imbalance, a common characteristic of educational activity data. NPGC preserves empirical category proportions under this imbalance while satisfying privacy constraints. We demonstrate that the NPGC supports practical data sharing for instructional analytics without exposing individual-level records. We implement NPGC in Python as a synthesizer wrapper designed for straightforward integration into existing data workflows, enabling practical plug-and-play deployment.\footnote{\url{https://github.com/gdiaz95/Synthetic_data_generation}}

This paper is organized as follows. Section~\ref{related_work} reviews prior approaches to tabular data generation and related instability phenomena. Section~\ref{Our_method} presents the formal formulation of NPGC. Section~\ref{experiments_explained} describes the benchmark datasets and the evaluation protocols. Section~\ref{results} reports comparative results on fidelity, privacy, and regeneration stability. Section~\ref{use_case} details the deployment of a real-world online platform use case. Sections~\ref{conclusion} and \ref{future} discuss, conclude, and outline future directions.

\section{Related Work}
\label{related_work}

Synthetic tabular data generation has long been studied as a mechanism for releasing useful data while controlling disclosure risk~\cite{Reiter2004,SurveyTabular}. In education, the tension between utility, fidelity, privacy, and validity constraints is particularly pronounced: Student-level datasets are high-dimensional and heterogeneous, combining continuous, integer/count, categorical, and ordinal variables within a single table, yet institutional and regulatory constraints often prevent broad sharing for benchmarking and reproducibility. Recent surveys and benchmark studies show that reported rankings of tabular generators can depend strongly on evaluation protocols, hyperparameter tuning effort, and available compute resources~\cite{Hansen2023,Lautrup2024,ExtensiveTuningBenchmark,stoian2024}. These findings motivate synthesis methods that are not only competitive in utility but also stable, reproducible, and computationally accessible.

Following the requirement-oriented taxonomy of \cite{SurveyTabular}, we summarize representative model families and position our approach relative to their design trade-offs.

\subsection{Deep Learning for Tabular Data}

Deep generative models are widely used for tabular synthesis to capture complex cross-feature dependencies. GAN-based approaches such as TGAN~\cite{xu2018ctgan}, CTGAN~\cite{Xu2019CTGAN} and TVAE~\cite{Synthcity}, and CTAB-GAN/CTAB-GAN+~\cite{Zhao2021,Zhao2022} adapt the adversarial framework of~\cite{Goodfellow2014} to mixed continuous and categorical data through conditional training and tailored normalization schemes. Deep learning models often achieve strong downstream predictive performance. However, because they optimize global adversarial or reconstruction objectives, mar\-ginal distributions are not explicitly constrained; univariate shapes can deviate from empirical distributions \cite{Kossale2022}, particularly for skewed or multi-modal educational attributes (Fig.~\ref{fig:problem_image}).

Diffusion- and score-based extensions, including TabDDPM \cite{Kotelnikov2023}, STaSy~\cite{Kim2023}, CoDi~\cite{Lee2023}, AutoDiff~\cite{Suh2023}, FinDiff~\cite{Sattarov2023}, and TabSyn~\cite{Zhang2024}, increase modeling flexibility through multi-step denoising objectives~\cite{SohlDickstein2015,Ho2020}. LLM-based generators such as GReaT~\cite{Borisov2023}, built on Transformer architectures~\cite{Vaswani2017}, provide flexible conditioning at the cost of substantial computational resources. Empirical studies further indicate that deep learning does not consistently outperform simpler approaches on tabular data and typically requires greater tuning effort~\cite{DL_is_not,ExtensiveTuningBenchmark}. 

Given our objective of providing a lightweight, stable, and reproducible synthesizer for privacy-sensitive educational da\-ta\-sets, we restrict our deep learning baselines to representative GAN/VAE models (CTGAN and TVAE), enabling controlled comparison against a statistical alternative.

\subsection{Parametric and Copula-Based Statistical Methods}

Copulas provide a classical framework for modeling multivariate dependence by separating marginals from the joint structure under Sklar’s theorem~\cite{SklarCopulas}. Gaussian and vine copulas are widely used in finance, marketing, survival analysis, and applied statistics~\cite{Song2000,Danaher2010,othus2010,Czado2010,Czado2022,Joe2014}. Their decoupled formulation enables efficient sampling and interpretability, motivating synthetic tabular systems such as the Gaussian-copula model in SDV~\cite{SDV} and related copula-based generators for complex data~\cite{Benali2021}.

Most copula-based synthesizers assume parametric marginal families. When educational attributes are skewed, discrete, or multi-modal, such assumptions can introduce systematic distortion. Differentially private extensions such as DPCopula~\cite{DPCopula} inject calibrated noise into marginal and correlation estimates to satisfy formal DP guarantees, but privacy noise may exacerbate misspecification effects. Nonparametric copula generators based on empirical CDFs relax distributional assumptions~\cite{Restrepo2023}, yet they do not explicitly address DP accounting, regeneration stability, or reproducible low-overhead benchmarking in privacy-restricted educational settings.

NPGC retains the copula decomposition while replacing pa\-rametric marginals with privatized empirical anchors and enforcing positive semi-definite correction of the noisy correlation matrix. This design prioritizes exact marginal fidelity, computational efficiency, and reproducibility under constrained compute, positioning NPGC as a lightweight statistical alternative rather than a high-capacity hybrid or deep generator.

\subsection{Recursive Training and Synthetic \\ Feedback Effects}
\label{MAD}

Researchers increasingly reuse synthetically generated data for downstream training, distillation, and iterative regeneration. When models train on data generated by earlier models, distributional drift can arise. Recent work documents instability under recursive self-training, including Model Autophagic Disorder (MAD)~\cite{Alemohammad2024}, model collapse~\cite{Shumailov2024}, degradation of diversity when scaling with synthetic data alone~\cite{Feng2024}, and progressive loss of variance and tail behavior in large models trained on recursively generated corpora~\cite{Shumailov2024}. In GAN settings, researchers attribute such behavior to the concentration of probability mass on limited modes and propose mitigation strategies such as VEEGAN~\cite{Srivastava2017}. Related analyses in LLM fine-tuning show that recursive adaptation can amplify distributional bias~\cite{Meng2024}.

Although this literature focuses primarily on vision and language models, the underlying mechanism—accumulation of approximation error under repeated self-consumption—ex\-tends directly to tabular synthesis. In tabular data, even small marginal distortions can compound across generations, attenuating rare categories or low-frequency subgroups. In educational settings, such drift is particularly consequential because subgroup underrepresentation may bias downstream analytics, fairness assessments, and policy decisions.

To the best of our knowledge, model collapse and regeneration stability have not been systematically analyzed in privacy-sensitive tabular synthesis. We therefore introduce an explicit empirical evaluation of synthetic feedback effects in tabular data. In Section~\ref{results}, we examine how marginal preservation influences recursive stability and whether constraining marginals mitigates variance shrinkage under synthetic feedback.

\section{Non-Parametric Gaussian \\  Copula (NPGC)}
\label{Our_method}
\gabriel{added text lines (not formulas) to each part of NPGC addressing reviewer 1 bullet 3}

We formulate NPGC as a three-stage generative process grounded in Sklar’s Theorem~\cite{SklarCopulas}, which decomposes a joint distribution into univariate marginal distributions and a co\-pu\-la-based dependency structure. To provide formal privacy guarantees, differential privacy (DP)~\cite{DP} is incorporated directly into both marginal estimation and correlation modeling. The procedure is defined as follows:

\subsection{Marginal Transformation: $X \to Z$}
\label{forward}
Intuitively, this step reorders each variable into a common Gaussian scale while storing its observed distribution, allowing dependencies to be modeled separately from marginal shapes. NPGC maps the original data matrix $X \in \mathbb{R}^{n \times p}$ into a latent Gaussian space $Z \in \mathbb{R}^{n \times p}$ column-wise, where $n$ denotes the number of records and $p$ the number of features. Privacy is enforced during marginal estimation prior to Gaussian projection. For each feature column $X_j$ ($j=1,\dots,p$), the transformation proceeds in two steps: 
\vspace{-1.1em}

\begin{enumerate}
     \item Mapping of $X_j$ to a variable $U_j \in (0,1)$ via a column-specific empirical probability integral transform; under consistency of $\widehat{F}_j$, $U_j \Rightarrow \mathcal{U}(0,1)$.
     \vspace{-0.2em}
    \item Projection to the standard Gaussian space via $Z_j = \Phi^{-1}(U_j)$, where $\Phi$ denotes the cumulative distribution function of the standard normal distribution.
\end{enumerate}
\vspace{-1.1em}

For continuous variables, we map observed values to the unit interval using a histogram-based estimate of the marginal CDF. Let $\widehat{F}_j$ denote the empirical CDF constructed from the non-missing entries of column $j$. We set $U_j=\widehat{F}_j(X_j)$ using linear interpolation, so that $U_j\in(0,1)$. 

For integer-valued variables with empirical probability mass function $p_j$ and CDF $\widehat{F}_j$, we apply dequantization $U_j=\widehat{F}_j(X_j)-r\,p_j(X_j)$ with $r\sim\mathcal{U}(0,1)$, which distributes probability mass uniformly over $(\widehat{F}_j(x^-),\widehat{F}_j(x)]$ for each observed value $x$, where $x^{-}$ denotes the immediate predecessor of $x$ in the ordered discrete support. 

Categorical variables are treated analogously by estimating empirical category probabilities and sampling uniformly within each category’s CDF interval. This embeds continuous, integer, and categorical variables into a common continuous domain $(0,1)$.

Let $f_{\mathrm{nan}}$ denote the empirical fraction of missing values in column $j$. Observed numeric values are scaled to occupy $(0,1-f_{\mathrm{nan}})$, while the upper interval $(1-f_{\mathrm{nan}},1)$ is reserved for missingness. This produces a disjoint partition of $(0,1)$ in which observed variability and missingness are encoded separately. This representation also preserves correlations between missingness and observed variables. As a result, missingness patterns can be reproduced when they are reflected in the data (e.g., missing values occurring more frequently within specific subgroups). However, if missingness depends on unobserved factors, these dependencies cannot be fully recovered. After uniformization, we apply the Gaussian projection $Z_j=\Phi^{-1}(U_j)$, where $\Phi$ is the standard normal CDF, yielding latent variables with approximately standard normal marginals for correlation estimation.

At this stage, differential privacy is applied to ensure that individual records do not significantly affect the learned distributions, thereby protecting individual-level information. To satisfy $\epsilon$-differential privacy, the total privacy budget is split as $\epsilon=\epsilon_m+\epsilon_c$ with $\epsilon_m=\epsilon_c=\epsilon/2$, where $\epsilon_m$ is allocated to marginal estimation and $\epsilon_c$ to correlation estimation. Marginal distributions are privatized using the Laplace mechanism under $\epsilon$-differential privacy, with $\epsilon=1$ by default and configurable by the user. For histogram bin counts $h_k$, we construct $\tilde{h}_k=\max(0,h_k+\eta_k)$ with $\eta_k\sim\mathrm{Lap}(1/\epsilon_m)$. For categorical or integer counts $c_\ell$, we construct $\tilde{c}_\ell=\max(0,c_\ell+\eta_\ell)$ with $\eta_\ell\sim\mathrm{Lap}(1/\epsilon_m)$. The empirical CDF $\widehat{F}_j$ is then formed from the normalized perturbed counts and stored for inverse sampling (Section~\ref{inverse}).

The resulting matrix $Z \in \mathbb{R}^{n \times p}$ has standard normal margi\-nals across columns. Specifically, let $Z_j$ denote the $j$-th column of $Z$. Then $Z_j \sim \mathcal{N}(0,1)$, and the columns are suitable for correlation estimation in the latent Gaussian space.

\subsection{Correlation and Gaussian Sampling}
\label{correlation}
In simple terms, this step captures how variables co-vary after removing their individual distributions, enabling joint dependence to be reproduced independently of individual form. Once the data is transformed into the Gaussian space $Z$, we estimate and replicate the dependency structure. We compute the empirical correlation matrix $\widehat{R} = \frac{1}{n - 1} Z^{\top}Z$. The matrix is privatized using the correlation budget $\epsilon_c$ by adding symmetric Laplace noise with scale $2/(n\epsilon_c)$ to its entries. Let $\tilde{R}$ denote the noisy matrix. To enforce positive semi-definiteness, we compute the eigen-decomposition $\tilde{R} = V \Lambda V^{\top}$ and replace $\Lambda$ with $\Lambda^+ = \operatorname{diag}(\max(\lambda_k, \delta))$, where we use $\delta = 10^{-8}$ in practice. The matrix is then reconstructed as $\widehat{R} = V \Lambda^+ V^{\top}$ and normalized to have a unit diagonal. The resulting $\widehat{R}$ is stored for subsequent sampling.

We generate $Z_{\mathrm{iid}} \in \mathbb{R}^{m\times p}$, where $m$ denotes the number of generated samples, with rows sampled independently as $\mathcal{N}(0, I_p)$. Let $\widehat{R}=LL^{\top}$ denote the Cholesky factorization of the learned correlation matrix. Correlated samples are obtained via $Z_{\text{syn}} = Z_{\mathrm{iid}}L^{\top}.$ Each row of $Z_{\text{syn}}$ then follows $\mathcal{N}(0,\widehat{R})$, yielding a synthetic Gaussian matrix that preserves the learned dependency structure.

\subsection{Marginal Reconstruction: $Z_{\text{syn}} \to X_{\text{syn}}$}
\label{inverse}

This step converts the synthetic data back into the original format of the dataset using the empirical marginal anchors, preserving the original distributions. Since sampling depends only on the privatized marginals $\widehat{F}_j$ and correlation matrix $\widehat{R}$, the inverse stage is a post-processing operation. We now map the Gaussian samples $Z_{\text{syn}}$ back to the original feature space to obtain $X_{\text{syn}}$ by reversing the forward transformation: 
\vspace{-1.15em}
\begin{enumerate}
    \item Map to uniform domain via the standard normal CDF: $U_{\text{syn},j} = \Phi(Z_{\text{syn},j})$.
    \vspace{-0.25em}
    \item Apply inverse empirical CDF, $X_{\text{syn},j} = \widehat{F}_j^{-1}(U_{\text{syn},j}),$ using interpolation for continuous variables and bin assignment for discrete types.
\end{enumerate}
\vspace{-1.15em}

The generative process yields $X_{\text{syn}} \in \mathbb{R}^{m \times p}$. Marginals are generated using empirical anchors stored during fitting: sorted empirical values for numeric columns and (noisy) category masses for categorical columns, applied through inverse empirical CDF sampling. In contrast to standard Gaussian copula models that assume parametric marginal families, NPGC relies directly on empirical (and privatized) marginal distributions while modeling dependencies solely through the Gaussian copula correlation structure.

To simplify usage, NPGC is implemented as a synthesizer class following a standard fit–sample paradigm. As shown in Listing~\ref{lst:npgc_usage}, the user fits the model to a tabular dataset. The fit step estimates the empirical marginal distributions and the Gaussian copula correlation matrix. After fitting, the model can generate synthetic samples of arbitrary size, and the trained instance can be saved to or loaded from disk for reproducible generation. In practice, this follows a straightforward workflow: fit the model to real data, then generate synthetic samples from the learned distributions.

\begin{lstlisting}[language=Python, caption={Example usage of the Non-Parametric Gaussian Copula (NPGC) synthesizer.\vspace{0.3em}},label={lst:npgc_usage}]
# Instantiate the NPGC synthesizer
synthesizer = NPGC()

# Fit the model to the original dataset (pandas DataFrame input)
synthesizer.fit(original_dataframe)

# Alternatively, load a previously trained model
synthesizer.load("models/synthetic_model.pkl")

# Generate synthetic samples (returned as a pandas DataFrame)
synthetic_data = synthesizer.sample(num_rows=1000)

# Save the fitted model for future reuse
synthesizer.save("models/synthetic_model.pkl")
\end{lstlisting}

\begin{figure}[t]
    \centering
    \Description{}
    \includegraphics[width=0.95\linewidth]{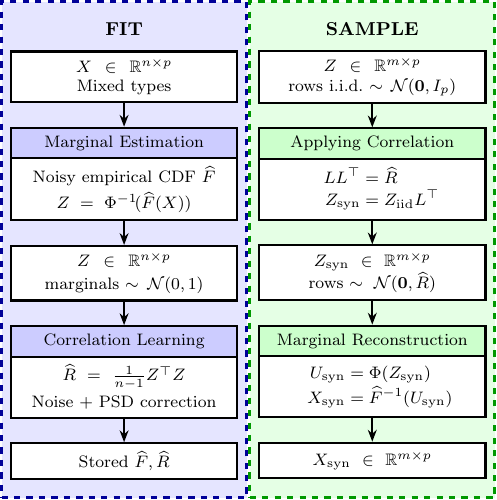}
    \caption{Overview of NPGC. FIT: Empirical marginal estimation via empirical Cumulative Distribution Functions (CDFs) with Gaussian projection, followed by correlation estimation $\widehat{R}$ with positive semi-definite (PSD) correction; the collection of marginals $\{\widehat{F}_j\}_{j=1}^p$, where $p$ denotes the number of features (shown as $\widehat{F}$ in the figure for clarity), and $\widehat{R}$ are stored. SAMPLE: Independent Gaussian samples are correlated using Cholesky factorization $LL^{\top}$ of $\widehat{R}$ and mapped back through inverse empirical CDF reconstruction.}
    \label{fig:flowchart}
\end{figure}

An overview of the NPGC fit and sample procedures is shown in Fig.~\ref{fig:flowchart}. The model consists of two components: privatized empirical marginals and a Gaussian copula correlation matrix. The fit phase estimates these quantities in a latent Gaussian space, and the sample phase generates correlated Gaussian draws followed by inverse marginal reconstruction to produce synthetic data. Details of the computational complexity for the fit and sample procedures are provided in Appendix~\ref{complexity}.


\section{Experimental Setup}
\label{experiments_explained}

The datasets reflect heterogeneous tabular structure, varying in sample size from $n=649$ to $n>45{,}000$ and in dimensionality from $p=4$ to $p=36$, with mixtures of continuous, discrete, and categorical variables and missing values in several cases. We do not impose a theoretical minimum sample size. Robustness to dataset size is assessed empirically across benchmarks ranging from $n=625$ to $n>45{,}000$ (Table~\ref{tab:datasets}). NPGC maintains stable performance across this range, including the smallest dataset ($n=649$), without dataset-specific tuning. For each dataset, we generate a synthetic dataset of size equal to the training partition ($m=0.8n$). Each dataset is split once into fixed 80/20 train–holdout partitions for fidelity and privacy evaluation. For utility, a separate 70/30 split is used for TSTR evaluation, with both partitions using a predefined random seed. Synthetic samples are generated from the training split only. The holdout set is never used for model fitting and remains fixed across regeneration iterations.

We evaluate NPGC against four baseline models implemen\-ted using the SDV single-table synthesizers~\cite{SDV}, selected to represent the main model classes in tabular data synthesis: parametric copula models (PGC)~\cite{SklarCopulas}, deep generative models (CTGAN, TVAE)~\cite{Xu2019CTGAN,CTGAN_TVAE}, and hybrid approaches (Copula-GAN). These methods provide a consistent basis for comparison within the SDV framework, covering the main modeling paradigms used in tabular synthesis and enabling evaluation across parametric, deep learning, and hybrid approaches. Future work may evaluate NPGC against more recent tabular synthesis methods. Experiments are conducted on the benchmark datasets summarized in Table~\ref{tab:datasets}, obtained from the UCI Machine Learning Repository~\cite{Uci}.
\begin{table}[ht]
\centering
\small
\caption{Benchmark datasets used for evaluation, showing the number of instances and features for each dataset.}
\vspace{0.3 em}
\label{tab:datasets}
\begin{tabular}{lrr}
\toprule
\textbf{Dataset} & \textbf{Instances ($n$)} & \textbf{Features ($p$)} \\ \midrule
Adult~\cite{Adult} & 48,842 & 14 \\
Balance Scale~\cite{Balance} & 625 & 4 \\
Nursery~\cite{Nursery} & 12,960 & 8 \\
Student Dropout~\cite{Dropout} & 4,424 & 36 \\
Student Performance~\cite{Performance} & 649 & 33 \\ \bottomrule
\end{tabular}
\end{table}

We evaluate synthetic data quality using three standard criteria—fidelity, utility, and privacy—following established practice~\cite{SurveyTabular}. In addition, we introduce a regeneration stability protocol that measures robustness under iterative synthetic feedback. 

Fidelity measures agreement between the statistical structure of real and synthetic data. It is quantified using the SDMetrics single-table Quality Report ~\cite{SDV}, summarized by the Overall Score, defined as the arithmetic mean of the report’s univariate and bivariate components: Column Shapes (univariate distribution agreement, including Kolmogorov–Smirnov comparisons for numerical variables) and Column Pair Trends (pairwise relationship preservation). These scores lie in $[0,1]$, with 1 indicating perfect agreement under the SDMetrics definitions.

Privacy refers to protecting sensitive information in the original dataset and preventing the synthetic output from enabling re-identification attacks (e.g., linkage, inference, or membership attacks). In this work, privacy is evaluated using Discriminator AUC and Distance to Closest Record (DCR) Share~\cite{SDV}. Discriminator AUC is computed by training a binary classifier solely for evaluation to distinguish real from synthetic samples and measuring the area under the receiver operating characteristic curve. DCR Share quantifies proximity between synthetic and real records based on nearest-neighbor distances. Both metrics are bounded in $[0,1]$, and values closer to 0.5 indicate lower separability and reduced memorization risk.

Utility refers to how fit for use the synthetic dataset is for the intended analytic task. In this work, utility is evaluated using the Train-on-Synthetic, Test-on-Real (TSTR) protocol together with runtime measurements. An XGBoost classifier~\cite{Xgboost} (fixed random seed, default hyperparameters) is trained on synthetic data and evaluated on real data using classification accuracy. We select XGBOOST as a baseline for TSTR due to its strong performance, robustness to heterogeneous feature types, and widespread use in educational data mining. Future work may evaluate utility across a broader set of predictive models. Accuracy is bounded in $[0,1]$, with higher values indicating better predictive performance. Performance is compared to a classifier trained on real data under the same split, and relative degradation (Accuracy Drop) is reported. Training time denotes synthesizer fitting time, and evaluation time denotes synthetic sample generation time.

Regeneration stability measures preservation of fidelity under repeated Synthetic Feedback Loop (SFL) iterations. It is evaluated using a 10-step synthetic feedback protocol on the Adult dataset~\cite{Adult}. At each iteration, the model is reinitialized and retrained on the synthetic dataset generated at the previous generation. Let $S^{(k)}$ denote the synthetic dataset generated at iteration $k$ under SFL. Regeneration stability is defined as the preservation of the Overall Score between $S^{(k)}$ and the original real dataset across $k=1,\dots,10$.

\section{Results}
\label{results}

We evaluate NPGC against four baseline models across the five benchmark datasets described in Section~\ref{experiments_explained}. Results are reported for fidelity, utility, privacy, and regeneration stability and are averaged across all datasets (one aggregate score per dataset).

\begin{figure}[t]
    \centering
    \Description{}
    \includegraphics[width=0.95\linewidth]{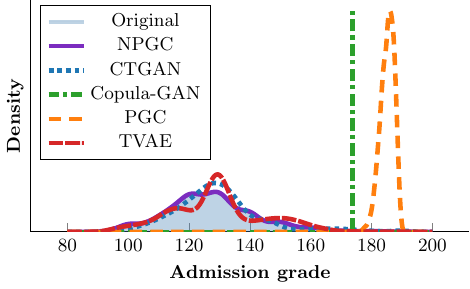}
    \caption{Marginal density of the Admission grade variable in the Student Dropout Success~\cite{Dropout} dataset. The shaded region denotes the empirical density of the real data. Curves correspond to TGAN, TVAE, PGC, Copula-GAN, and the proposed NPGC. Deep learning and parametric baselines distort the marginal shape, while NPGC preserves it through empirical anchoring.}
    \label{fig:solution_image}
\end{figure}

\subsection{Fidelity}
\gabriel{added text lines (not results) to each part of results addressing reviewer 1 bullet 3}
Fidelity measures how closely the synthetic data matches the statistical structure of the real data. In practice, higher fidelity indicates that summary statistics and distributions remain consistent between real and synthetic datasets. As shown in Table~\ref{tab:results_avg}, NPGC achieves the highest Overall Score (0.962), as well as the highest Column Shapes score (0.985) and Column Pair Trends score (0.939). These results indicate that NPGC consistently outperforms all baselines in fidelity across both univariate and pairwise correlation preservation.

\begin{table}[hb]
\centering
\small
\caption{Average fidelity metrics across datasets. Overall Score reflects univariate (Column Shapes) and bivariate (Pair Trends) agreement between synthetic and real data. NPGC attains the highest fidelity across all metrics.}
\vspace{0.3em}
\label{tab:results_avg}
\begin{tabular}{lccc}
\toprule
\textbf{Model} & \textbf{Overall Score} & \textbf{Column Shapes} & \textbf{Pair Trends} \\ \midrule
\textbf{NPGC} & \textbf{0.962} & \textbf{0.985} & \textbf{0.939} \\ 
CTGAN & 0.895 & 0.918 & 0.872 \\
Copula-GAN & 0.888 & 0.914 & 0.863 \\
PGC & 0.897 & 0.913 & 0.881 \\
TVAE & 0.832 & 0.883 & 0.781 \\ 
\bottomrule
\end{tabular}
\end{table}

Figure~\ref{fig:solution_image} compares the marginal distribution of the \textit{Admission grade} variable in the Student Dropout Success~\cite{Dropout} dataset. The Copula-GAN collapses the distribution into a narrow spike near $x\approx177.8$, indicating incorrect assumptions. PGC produces a unimodal approximation that does not capture the non-standard structure of the real data. NPGC consistently reproduces the empirical marginal.

\subsection{Privacy}
\label{privacy_results}
Privacy measures the extent to which individual records from the real dataset are protected. In practice, stronger privacy indicates lower risk of distinguishing or linking synthetic samples to real individuals. As shown in Table~\ref{tab:privacy_table}, NPGC achieves the lowest Discriminator AUC (0.801) and DCR Share (0.527). Since values closer to 0.5 indicate lower separability and memorization, NPGC provides the strong\-est privacy performance among all methods.

\begin{table}[hb]
\centering
\small
\caption{Average privacy metrics across datasets. Lower values indicate reduced real–synthetic separability and lower memorization risk. NPGC achieves the strongest privacy performance.}
\vspace{0.3em}
\label{tab:privacy_table}
\begin{tabular}{lcc}
\toprule
\textbf{Model} & \textbf{Discriminator AUC} & \textbf{DCR Share} \\ \midrule
\textbf{NPGC} & \textbf{0.801} & \textbf{0.527} \\
CTGAN & 0.853 & 0.567 \\
Copula-GAN & 0.871 & 0.558 \\
PGC & 0.850 & 0.549 \\
TVAE & 0.910 & 0.545 \\ 
 \bottomrule
\end{tabular}
\end{table}

\subsection{Utility}
Utility measures how useful the synthetic data is for downstream tasks. In practice, this reflects whether models train\-ed on synthetic data behave similarly when applied to real data. Tables~\ref{tab:utility_table} and~\ref{tab:computational_time} show that TVAE attains the lowest accuracy drop (8.66\%), while NPGC achieves a competitive drop of 16.99\%, outperforming CTGAN and Copula-GAN and providing stronger privacy guarantees than TVAE.

\begin{figure}[t]
     \centering
     \Description{}
     \includegraphics[width=0.95\linewidth]{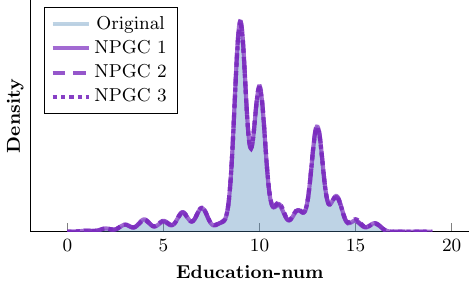}
    \caption{Marginal density of the \textit{Education-num} variable in the Adult~\cite{Adult} dataset across three sequential regeneration iterations of NPGC. The shaded region denotes the empirical density of the original data, and purple curves (iterations 1–3) represent synthetic samples generated under repeated regeneration. The overlap across iterations demonstrates stability of the marginal distribution.}
     \label{fig:mode_collapse}
\end{figure}

\begin{table}[hbt]
\centering
\small
\caption{Average Train-on-Synthetic, Test-on-Real (TSTR) utility metrics across datasets using XGBoost. Synthetic Score denotes classification accuracy when training on synthetic data and testing on real data, and Accuracy Drop measures the relative decrease compared to training on real data. NPGC achieves the second-highest utility.}
\vspace{0.3em}
\label{tab:utility_table}
\begin{tabular}{lcc}
\toprule
\textbf{Model} & \textbf{Synthetic Score} & \textbf{Accuracy Drop (\%)} \\ \midrule
\textbf{NPGC} & \textbf{0.739} & \textbf{16.99} \\
CTGAN & 0.619 & 25.73 \\
Copula-GAN & 0.647 & 22.80 \\
PGC & 0.675 & 17.04 \\
TVAE & 0.797 & 8.66 \\ 
 \bottomrule
\end{tabular}
\end{table}

Training times differ substantially: CTGAN and Copula-GAN exceed 790 seconds, TVAE requires 268 seconds, PGC 11.86 seconds, and NPGC 0.65 seconds. Sampling times are below 1.2 seconds for all methods. NPGC, as a non-deep-learning model, achieves orders-of-magnitude speedups over deep learning baselines and remains substantially faster than PGC baseline.

\begin{table}[hbt]
\centering
\small
\caption{Average computational runtimes across datasets. Train Time denotes model fitting time, and Eval Time denotes sampling and evaluation time. NPGC achieves the lowest training time.}
\vspace{0.3em}
\label{tab:computational_time}
\begin{tabular}{lcc}
\toprule
\textbf{Model} & \textbf{Train Time (s)} & \textbf{Eval Time (s)} \\ \midrule
\textbf{NPGC} & \textbf{0.650} & \textbf{1.19} \\ 
CTGAN & 821.01 & 0.68 \\
Copula-GAN & 791.90 & 0.75 \\
PGC & 11.860 & 1.03 \\
TVAE & 268.46 & 0.34 \\
\bottomrule
\end{tabular}
\end{table}

\subsection{Regeneration Stability}
Regeneration stability measures how well the synthetic data preserves its fidelity under repeated generation. In practice, this indicates whether the data maintains its variability and distribution across iterations. Regeneration stability is evaluated only on the Adult~\cite{Adult} dataset using the 10-step synthetic feedback protocol described in Section~\ref{experiments_explained}. As shown in Fig.~\ref{fig:mode_collapse}, we observe the marginal distribution of the \textit{Education-num} variable across successive NPGC iterations. NPGC correctly preserves the multi-modal empirical marginal distribution across iterations with minimal drift by construction, illustrating the absence of model collapse under repeated regeneration.

\begin{figure*}[t]
    \centering
    \Description{Line plot of Overall Score over 10 iterations. The NPGC line (purple) remains stable near 96 percent, while TVAE (red) and CTGAN (blue) show lower scores and more fluctuations.}
    \includegraphics[width=0.95\linewidth]{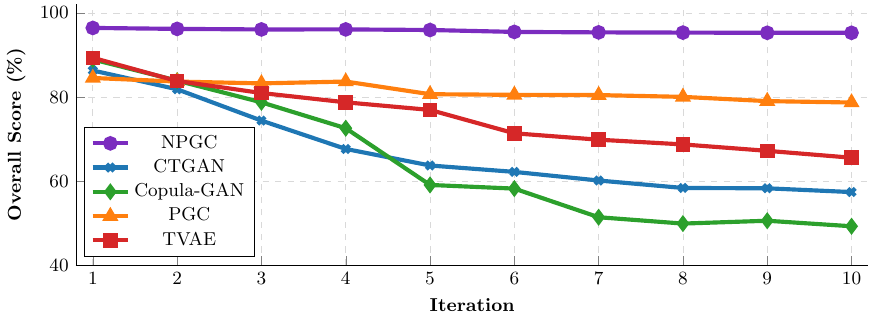}
    \caption{Overall Score (\%) across 10 synthetic regeneration iterations on the Adult~\cite{Adult} dataset. The Overall Score aggregates univariate and bivariate distributional agreement between synthetic and real data. Curves correspond to CTGAN, Copula-GAN, PGC, TVAE, and NPGC.}
    \label{fig:iterations}
\end{figure*}

This marginal stability is reflected in the aggregate fidelity metric. As shown in Fig.~\ref{fig:iterations}, NPGC Overall Score decreases from 96.43\% at iteration 1 to 95.27\% at iteration 10 (1.16 percentage points). Copula-GAN decreases from 88.93\% to 49.25\%, CTGAN from 86.31\% to 57.38\%, TVAE from 89.33\% to 65.56\%, and PGC from 84.53\% to 78.69\%. These results show that NPGC maintains marginal and bivariate structure (as captured by the Overall Score) under repeated regeneration, while alternative methods exhibit substantial degradation.

\begin{figure}[t]
     \centering
     \Description{}
    \includegraphics[width=0.95\linewidth]{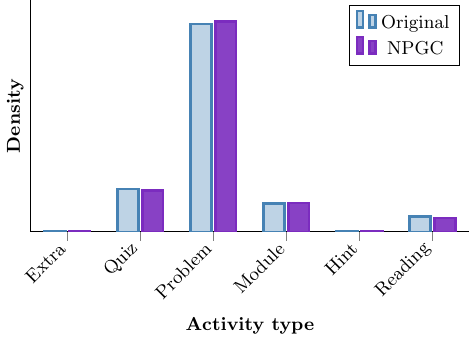}
    \caption{Proportion of activity types in the real and synthetic interaction dataset. The shaded bars represent empirical proportions from 7{,}506 real records. Synthetic proportions from 10{,}332 samples generated with NPGC closely match the empirical distribution, including low-frequency categories such as Hint and Extra. NPGC illustrates stable marginal preservation under deployment.}
     \label{fig:use_case}
\end{figure}

\section{Educational Research \\Application}
\label{use_case}

In addition to benchmark datasets, we evaluate NPGC in an applied setting using interaction logs from a large-scale online learning platform. The dataset contains 7{,}506 xAPI interaction records with an \textit{activity\_type} categorical variable comprising six levels: Module, Reading, Quiz, Problem, Extra, and Hint (see Appendix~\ref{app:platform} for category definitions). The distribution is highly imbalanced: Problem events account for more than 70\% of interactions, while Hint events occur in fewer than 0.2\% of records.

The objective in this applied scenario is to enable the release of augmented synthetic data modeled after the underlying data while preserving exact categorical proportions required by researchers interested in downstream analysis of the data. With NPGC, we generate an augmented synthetic dataset of size 10{,}332 samples. Due to platform privacy policies, the raw interaction data cannot be publicly released; however, SDG follows the same protocol as previously described. We compare activity-type proportions between the real and synthetic datasets. As shown in Fig.~\ref{fig:use_case}, NPGC closely matches the empirical distribution across all six categories, including extreme minority classes. The synthetic dataset contains 7{,}227 Problem events, 1443 Quiz events, 1067 Module events, 552 Reading events, 27 Extra events, and 16 Hint events, preserving the observed imbalance structure.

We show that NPGC preserves categorical proportions under severe class imbalance, supporting privacy-preserving data release and augmentation without altering event frequencies.

\section{Discussion}
\label{conclusion}

Our results carry several implications for synthetic data practice in educational research. We organize this discussion around three themes: what NPGC's marginal preservation means for analytic validity, what its limitations imply for specific research designs, and what directions would most benefit the EDM community.

\textbf{Safeguarding Rare Educational Signals.} When synthesizers smooth multi-modal distributions or collapse low-frequency categories, as Copula-GAN and CTGAN did across benchmarks, the distortion is not evenly distributed. Rare subgroups are disproportionately affected: sparsely observed demographic categories, low-frequency event types, and tail behaviors are attenuated first. In educational datasets, these are often the populations of greatest analytic interest: underrepresented students, rare help-seeking behaviors, or a\-typical learning trajectories. NPGC's empirical anchoring preserves these distributional features by construction, reducing the risk that synthetic data silently erases the students researchers most need to study. Our deployment (Section~\ref{use_case}) illustrates this concretely: NPGC maintained the exact categorical proportions of a severely imbalanced activity-type distribution, including Hint events occurring in fewer than 0.2\% of records.

\textbf{The utility-fidelity tradeoff.} NPGC achieves the highest fidelity but not the highest downstream classification accuracy; TVAE's accuracy drop (8.66\%) is lower than NPGC's (16.99\%). This gap likely reflects a fundamental tension: strict marginal anchoring constrains the joint distribution in ways that can sacrifice cross-feature predictive signal, particularly nonlinear interactions that deep generators capture through flexible function approximation. However, for many EDM use cases such as population characterization, distributional auditing, subgroup representation analysis, and descriptive reporting, marginal fidelity matters more than predictive utility. These differences have direct implications for common EDM tasks. In descriptive analyses, higher fidelity means that distributions such as grades, activity counts, or demographic proportions remain consistent, so reported statistics do not change when using synthetic data. In predictive settings, comparable TSTR performance means that models trained on synthetic data behave similarly when evaluated on real data. For tasks such as dropout prediction or early warning systems, this reduces the risk that model outputs or intervention thresholds shift due to artifacts introduced during synthesis. Researchers should select their synthesizer based on their analytic objective. 

\textbf{Limitations and scope conditions.} The Gaussian copula captures linear dependence structure in the latent space. Educational data frequently exhibits threshold effects, ceiling and floor artifacts, and nonlinear interactions (e.g., between prior knowledge and intervention response) that this assumption may not preserve. Additionally, our regeneration stability analysis was conducted on a single dataset; whether these properties generalize to high-cardinality educational features (e.g., course IDs or learning objective codes) remains an open question.

\section{Future Directions}
\label{future}
The most impactful extensions would address structures co\-mmon in educational research: longitudinal panel data with temporal dependencies, multi-table relational schemas (students × courses × assessments), and conditional distributions relevant to causal inference in platform A/B experiments. Integrating flexible dependency models, such as vine copulas~\cite{Czado2022} or neural spline flows~\cite{Splines}, while retaining empirical marginal anchoring could extend NPGC's applicability to these richer data structures without sacrificing its core marginal stability and reproducibility guarantees. Future work may also provide a more precise characterization of the dependency structures captured by the model, evaluate NPGC against more recent tabular synthesis methods, and assess utility across a broader range of predictive models.

%
\bibliographystyle{abbrv}
\bibliography{references}  

@string{neurips = "Advances in Neural Information Processing Systems"}

@string{edm = "International Conference on Educational Data Mining"}

@string{iclr = "International Conference on Learning Representations"}

@string{icml = "International Conference on Machine Learning"}

@string{acml = "Asian Conference on Machine Learning"}

@string{icaif = "International Conference on AI in Finance"}

@string{dsaa = "International Conference on Data Science and Advanced Analytics"}

@string{icoa = "International Conference on Optimization and Applications"}

@string{kdd = "International Conference on Knowledge Discovery and Data Mining"}

@string{cedt = "International Conference on Extending Database Technology"}

@string{lakc = "International Learning Analytics and Knowledge Conference"}

@article{pollution,
  title   = {Synthetic data: From data scarcity to data pollution},
  author  = {Wiehn, Tanja},
  journal = {Surveillance \& Society},
  year    = {2024},
  volume  = {22},
  number  = {4},
  pages   = {472--476},
  doi     = {10.24908/ss.v22i4.18327}
}

@incollection{DP,
  title     = {Calibrating noise to sensitivity in private data analysis},
  author    = {Dwork, Cynthia and McSherry, Frank and Nissim, Kobbi and Smith, Adam},
  booktitle = {Theory of Cryptography},
  publisher = {Springer},
  year      = {2006},
  pages     = {265--284},
  doi       = {10.1007/11681878_14}
}

@misc{Uci,
  author       = {Dua, Dheeru and Graff, Casey},
  title        = {UCI Machine Learning Repository},
  year         = {2017},
  note         = {University of California, Irvine, School of Information and Computer Sciences}
}

@misc{Adult,
  title  = {{Adult}},
  author = {Becker, Barry and Kohavi, Ronny},
  howpublished = {UCI Machine Learning Repository},
  year   = {1996},
  doi    = {10.24432/C5XW20}
}

@misc{Balance,
  title  = {{Balance Scale}},
  author = {Siegler, Robert},
  howpublished = {UCI Machine Learning Repository},
  year   = {1976},
  doi    = {10.24432/C5488X}
}

@misc{Nursery,
  title  = {{Nursery}},
  author = {Olave, Manuel and Rajkovic, Vladislav and Bohanec, Marko},
  howpublished = {UCI Machine Learning Repository},
  year   = {1989},
  doi    = {10.24432/C5P88W}
}

@InProceedings{Dropout,
  author    = {Martins, M{\'o}nica V. and Tolledo, Daniel and Machado, Jorge and Baptista, Lu{\'i}s M. T. and Realinho, Valentim},
  title     = {Early Prediction of Student's Performance in Higher Education: A Case Study},
  booktitle = {Trends and Applications in Information Systems and Technologies},
  year      = {2021},
  pages     = {166--175}
}

@misc{Performance,
  author       = {Cortez, Paulo},
  title        = {{Student Performance}},
  year         = {2008},
  howpublished = {{UCI Machine Learning Repository}},
  doi          = {10.24432/C5TG7T}
}

@inproceedings{Xgboost,
  title     = {{XGBoost}: A scalable tree boosting system},
  author    = {Chen, Tianqi and Guestrin, Carlos},
  booktitle = kdd,
  year      = {2016},
  pages     = {785--794}
}

@inproceedings{Splines,
  title     = {Neural spline flows},
  author    = {Durkan, Conor and Bekasov, Artur and Murray, Iain and Papamakarios, George},
  booktitle = neurips,
  year      = {2019}
}

@article{Synthcity,
  title   = {{Synthcity}: Facilitating innovative use cases of synthetic data in different data modalities},
  author  = {Qian, Zhaozhi and Cebere, Bogdan-Constantin and van der Schaar, Mihaela},
  journal = {\arxiv{2301.07573}},
  year    = {2023}
}

@article{SklarCopulas,
  author  = {Sklar, A.},
  title   = {Fonctions de r{\'e}partition {\`a} n dimensions et leurs marges},
  journal = {Publications de l'Institut de Statistique de l'Universit{\'e} de Paris},
  volume  = 8,
  pages   = {229--231},
  year    = 1959
}

@inproceedings{CTGAN_TVAE,
  author    = {Xu, L. and Skoularidou, M. and Cuesta{-}Infante, A. and Veeramachaneni, K.},
  title     = {Modeling tabular data using conditional {GAN}},
  booktitle = neurips,
  volume    = 32,
  year      = 2019
}

@article{SurveyTabular,
  author  = {Stoian, M. C. and Giunchiglia, E. and Lukasiewicz, T.},
  title   = {A survey on tabular data generation: Utility, alignment, fidelity, privacy, and beyond},
  journal = {\arxiv{2503.05954}},
  year    = 2025
}

@article{privacy2012,
  author  = {Goldfarb, A. and Tucker, C.},
  title   = {Privacy and innovation},
  journal = {Innovation Policy and the Economy},
  volume  = 12,
  pages   = {65--90},
  year    = 2012,
  doi     = {10.1086/663156}
}

@inproceedings{Klose_undated-dp,
  title     = {{EDM} and privacy: Ethics and legalities of data collection, usage, and storage},
  author    = {Klose, Mark and Desai, Vasvi and Song, Yang and Gehringer, Edward},
  booktitle = edm,
  year      = {2020}
}

@article{Reiter2004,
  title   = {Releasing multiply imputed, synthetic public use microdata: An illustration and empirical study},
  author  = {Reiter, Jerome P.},
  journal = {Journal of the Royal Statistical Society: Series A (Statistics in Society)},
  year    = {2004},
  pages   = {185--205},
  doi     = {10.1111/j.1467-985X.2004.00343.x}
}

@inproceedings{hansen2023,
  title     = {Reimagining synthetic tabular data generation through data-centric {AI}: A comprehensive benchmark},
  author    = {Hansen, Lasse and Seedat, Nabeel and van der Schaar, Mihaela and Petrovic, Andrija},
  booktitle = neurips,
  year      = {2023},
  pages     = {33781--33823}
}

@article{Lautrup2024,
  title   = {Systematic review of generative modelling tools and utility metrics for fully synthetic tabular data},
  author  = {Lautrup, Anton Danholt and Hyrup, Tobias and Zimek, Arthur and Schneider-Kamp, Peter},
  journal = {ACM Computing Surveys},
  year    = {2024},
  doi     = {10.1145/3704437}
}

@article{ExtensiveTuningBenchmark,
  title   = {Tabular data generation models: An in-depth survey and performance benchmarks with extensive tuning},
  author  = {Kindji, G. Charbel N. and Rojas-Barahona, Lina M. and Fromont, Elisa and Urvoy, Tanguy},
  journal = {Neurocomputing},
  year    = {2025},
  doi     = {10.1016/j.neucom.2025.131655}
}

@inproceedings{stoian2024,
  title     = {How realistic is your synthetic data? Constraining deep generative models for tabular data},
  author    = {Stoian, Mihaela Catalina and Dyrmishi, Salijona and Cordy, Maxime and Lukasiewicz, Thomas and Giunchiglia, Eleonora},
  booktitle = iclr,
  year      = {2024}
}

@article{DL_is_not,
  title   = {Tabular data: Deep learning is not all you need},
  author  = {Shwartz-Ziv, Ravid and Armon, Amitai},
  journal = {Information Fusion},
  year    = {2022},
  pages   = {84--90},
  doi     = {10.1016/j.inffus.2021.11.011}
}

@article{xu2018ctgan,
  title   = {Synthesizing tabular data using generative adversarial networks},
  author  = {Xu, Lei and Veeramachaneni, Kalyan},
  journal = {\arxiv{1811.11264}},
  year    = {2018}
}

@inproceedings{Xu2019CTGAN,
  title     = {Modeling tabular data using conditional {GAN}},
  author    = {Xu, Lei and Skoularidou, Maria and Cuesta-Infante, Alfredo and Veeramachaneni, Kalyan},
  booktitle = neurips,
  year      = {2019}
}

@inproceedings{Zhao2021,
  title     = {{CTAB}-{GAN}: Effective table data synthesizing},
  author    = {Zhao, Zilong and Kunar, Aditya and Birke, Robert and Chen, Lydia Y.},
  booktitle = acml,
  year      = {2021},
  pages     = {97--112}
}

@article{Zhao2022,
  title   = {{CTAB}-{GAN}+: Enhancing tabular data synthesis},
  author  = {Zhao, Zilong and Kunar, Aditya and Birke, Robert and Van der Scheer, H. and Chen, Lydia Y.},
  journal = {Frontiers in Big Data},
  year    = {2024},
  doi     = {10.3389/fdata.2023.1296508}
}

@inproceedings{Goodfellow2014,
  title     = {Generative adversarial nets},
  author    = {Goodfellow, Ian J. and Pouget-Abadie, Jean and Mirza, Mehdi and Xu, Bing and Warde-Farley, David and Ozair, Sherjil and Courville, Aaron and Bengio, Yoshua},
  booktitle = neurips,
  year      = {2014}
}

@inproceedings{Kotelnikov2023,
  title     = {{TabDDPM}: Modelling tabular data with diffusion models},
  author    = {Kotelnikov, Akim and Baranchuk, Dmitry and Rubachev, Ivan and Babenko, Artem},
  booktitle = icml,
  year      = {2023}
}

@inproceedings{Lee2023,
  title     = {{CoDi}: Co-evolving contrastive diffusion models for mixed-type tabular synthesis},
  author    = {Lee, Chaejeong and Kim, Jayoung and Park, Noseong},
  booktitle = icml,
  year      = {2023}
}

@inproceedings{Kim2023,
  title     = {{STaSy}: Score-based tabular data synthesis},
  author    = {Kim, Jayoung and Lee, Chaejeong and Park, Noseong},
  booktitle = iclr,
  year      = {2023}
}

@article{Suh2023,
  title   = {{AutoDiff}: Combining auto-encoder and diffusion model for tabular data synthesizing},
  author  = {Suh, Namjoon and Lin, Xiaofeng and Hsieh, Din-Yin and Honarkhah, Merhdad and Cheng, Guang},
  journal = {\arxiv{2310.15479}},
  year    = {2023}
}

@inproceedings{Sattarov2023,
  title     = {{FinDiff}: Diffusion models for financial tabular data generation},
  author    = {Sattarov, Timur and Schreyer, Marco and Borth, Damian},
  booktitle = icaif,
  year      = {2023},
  pages     = {64--72},
  doi       = {10.1145/3604237.3626876}
}

@inproceedings{Zhang2024,
  title     = {Mixed-type tabular data synthesis with score-based diffusion in latent space},
  author    = {Zhang, Hengrui and Zhang, Jiani and Shen, Zhengyuan and Srinivasan, Balasubramaniam and Qin, Xiao and Faloutsos, Christos and Rangwala, Huzefa and Karypis, George},
  booktitle = iclr,
  year      = {2024}
}

@inproceedings{SohlDickstein2015,
  title     = {Deep unsupervised learning using nonequilibrium thermodynamics},
  author    = {Sohl-Dickstein, Jascha and Weiss, Eric and Maheswaranathan, Niru and Ganguli, Surya},
  booktitle = icml,
  year      = {2015}
}

@inproceedings{Ho2020,
  title     = {Denoising diffusion probabilistic models},
  author    = {Ho, Jonathan and Jain, Ajay and Abbeel, Pieter},
  booktitle = neurips,
  year      = {2020}
}

@inproceedings{Borisov2023,
  title     = {Language models are realistic tabular data generators},
  author    = {Borisov, Vadim and Sessler, Kathrin and Leemann, Tobias and Pawelczyk, Martin and Kasneci, Gjergji},
  booktitle = iclr,
  year      = {2023}
}

@inproceedings{Vaswani2017,
  title     = {Attention is all you need},
  author    = {Vaswani, Ashish and Shazeer, Noam and Parmar, Niki and Uszkoreit, Jakob and Jones, Llion and Gomez, Aidan N. and Kaiser, {\L}ukasz and Polosukhin, Illia},
  booktitle = neurips,
  year      = {2017}
}

@inproceedings{DPCopula,
  title     = {Differentially private synthesization of multi-dimensional data using copula functions},
  author    = {Li, Huan and Xiong, Li and Jiang, Xiaoqian},
  booktitle = cedt,
  year      = {2014},
  pages     = {475--486},
  doi       = {10.5441/002/edbt.2014.43}
}

@inproceedings{SDV,
  title     = {The synthetic data vault},
  author    = {Patki, Neha and Wedge, Roy and Veeramachaneni, Kalyan},
  booktitle = dsaa,
  year      = {2016},
  pages     = {399--410},
  doi       = {10.1109/DSAA.2016.49}
}

@article{Restrepo2023,
  title   = {Nonparametric generation of synthetic data using copulas},
  author  = {Restrepo, J. P. and Rivera, J. C. and Laniado, H. and Osorio, P. and Becerra, O. A.},
  journal = {Electronics},
  year    = {2023},
  pages   = {1601},
  doi     = {10.3390/electronics12071601}
}

@book{Joe2014,
  title     = {Dependence modeling with copulas},
  author    = {Joe, Harry},
  publisher = {Chapman and Hall/CRC},
  year      = {2014},
  doi       = {10.1201/b17116}
}

@article{Song2000,
  title   = {Multivariate dispersion models generated from {Gaussian} copula},
  author  = {Song, Xue-Kun Peter},
  journal = {Scandinavian Journal of Statistics},
  year    = {2000},
  pages   = {305--320},
  doi     = {10.1111/1467-9469.00191}
}

@incollection{Czado2010,
  title     = {Pair-copula constructions of multivariate copulas},
  author    = {Czado, Claudia},
  booktitle = {Copula Theory and Its Applications},
  year      = {2010},
  publisher = {Springer},
  pages     = {93--109}
}

@article{Czado2022,
  title   = {Vine copula based modeling},
  author  = {Czado, Claudia and Nagler, Thomas},
  journal = {Annual Review of Statistics and Its Application},
  year    = {2022},
  pages   = {453--477},
  doi     = {10.1146/annurev-statistics-040220-101153}
}

@article{Danaher2010,
  title   = {Modeling multivariate distributions using copulas: Applications in marketing},
  author  = {Danaher, Peter J. and Smith, Michael S.},
  journal = {Marketing Science},
  year    = {2010},
  pages   = {4--21},
  doi     = {10.1287/mksc.1090.0491}
}

@inproceedings{Benali2021,
  title     = {{MTCopula}: Synthetic complex data generation using copula},
  author    = {Benali, Fodil and Bod{\'e}n{\`e}s, Damien and Labroche, Nicolas and de Runz, Cyril},
  booktitle = {International Workshop on Design, Optimization, Languages and Analytical Processing of Big Data (DOLAP)},
  year      = {2021},
  pages     = {51--60}
}

@article{othus2010,
  title   = {A {Gaussian} copula model for multivariate survival data},
  author  = {Othus, Megan and Li, Yi},
  journal = {Statistics in Biosciences},
  year    = {2010},
  pages   = {154--179},
  doi     = {10.1007/s12561-010-9026-x}
}

@inproceedings{Alemohammad2024,
  title     = {Self-consuming generative models go {MAD}},
  author    = {Alemohammad, Sina and Casco-Rodriguez, Josue and Luzi, Lorenzo and Humayun, Ahmed Imtiaz and Babaei, Hossein and LeJeune, Daniel and Siahkoohi, Ali and Baraniuk, Richard},
  booktitle = iclr,
  year      = {2024}
}

@inproceedings{Kossale2022,
  title     = {Mode collapse in generative adversarial networks: An overview},
  author    = {Kossale, Youssef and Airaj, Mohammed and Darouichi, Aziz},
  booktitle = icoa,
  year      = {2022},
  pages     = {1--6},
  doi       = {10.1109/ICOA55659.2022.9934291}
}

@article{Shumailov2024,
  title   = {{AI} models collapse when trained on recursively generated data},
  author  = {Shumailov, Ilia and Shumaylov, Zakhar and Zhao, Yiren and Papernot, Nicolas and Anderson, Ross and Gal, Yarin},
  journal = {Nature},
  year    = {2024},
  pages   = {755--759},
  doi     = {10.1038/s41586-024-07566-y}
}

@article{Feng2024,
  title   = {Beyond model collapse: Scaling up with synthesized data requires reinforcement},
  author  = {Feng, Yunzhen and Dohmatob, Elvis and Yang, Pu and Charton, Francois and Kempe, Julia},
  journal = {\arxiv{2406.07515}},
  year    = {2024}
}

@article{Meng2024,
  title     = {Attribute Controlled Fine-tuning for Large Language Models: A Case Study on Detoxification},
  author    = {Meng, Tao and Mehrabi, Ninareh and Goyal, Palash and Ramakrishna, Anil and Galstyan, Aram and Zemel, Richard and Chang, Kai-Wei and Gupta, Rahul and Peris, Charith},
  year      = {2024},
  journal = {Association for Computational Linguistics},
  pages     = {13329--13341},
  doi       = {10.18653/v1/2024.findings-emnlp.779}
}

@inproceedings{Srivastava2017,
  title     = {{VEEGAN}: Reducing mode collapse in {GAN}s using implicit variational learning},
  author    = {Srivastava, Akash and Valkov, Lazar and Russell, Chris and Gutmann, Michael U. and Sutton, Charles},
  booktitle = neurips,
  year      = {2017},
  volume    = {30}
}

@inproceedings{2023.EDM-long-papers.10,
  title     = {How to open science: Debugging reproducibility within the {Educational Data Mining} conference},
  author    = {Haim, Aaron and Gyurcsan, Robert and Baxter, Chris and Shaw, Stacy T. and Heffernan, Neil T.},
  booktitle = edm,
  year      = {2023},
  pages     = {114--124},
  doi       = {10.5281/zenodo.8115651}
}

@article{Shi2026LLMEducationReview,
  title   = {Large language models in education: A systematic review of empirical applications, benefits, and challenges},
  author  = {Shi, Yuhong and Yu, Kun and Dong, Yifei and Chen, Fang},
  journal = {Computers and Education: Artificial Intelligence},
  year    = {2026},
  doi     = {10.1016/j.caeai.2025.100529}
}

@inproceedings{Haim2023-qm,
  title     = {How to open science: A principle and reproducibility review of the learning analytics and knowledge conference},
  author    = {Haim, Aaron and Shaw, Stacy and Heffernan, Neil},
  booktitle = lakc,
  year      = {2023},
  pages     = {156--164},
  doi       = {10.1145/3576050.3576071}
}

@inproceedings{Phung2024-lz,
  title     = {Automating human tutor-style programming feedback: Leveraging {GPT-4} tutor model for hint generation and {GPT-3.5} student model for hint validation},
  author    = {Phung, Tung and P{\u{a}}durean, Victor-Alexandru and Singh, Anjali and Brooks, Christopher and Cambronero, Jos{\'e} Pablo and Gulwani, Sumit and Singla, Adish Kumar and Soares, Gustavo},
  booktitle = lakc,
  year      = {2024},
  pages     = {12--23}
}

@incollection{Du2025-jc,
  title     = {Privacy in fine-tuning large language models: Attacks, defenses, and future directions},
  author    = {Du, Hao and Liu, Shang and Zheng, Lele and Cao, Yang and Nakamura, Atsuyoshi and Chen, Lei},
  booktitle = {Advances in Knowledge Discovery and Data Mining},
  publisher = {Springer},
  year      = {2025},
  pages     = {326--344}
}
%
\appendix
\section{Online Platform Details}
\label{app:platform}

This appendix provides additional context for the anonymized online learning platform used in the case study described in Section~\ref{use_case}. Specific institutional names are withheld for peer review.

The dataset originates from a large-scale open educational resource (OER) platform that delivers digital instructional content and assessments to post-secondary learners. Student interactions are recorded using the Experience API (xAPI) standard, which logs fine-grained learning events in structured statement format. Each record captures a timestamped action performed by a learner within a course module.

The dataset used in this study consists of 7{,}506 interaction records. The primary variable of interest is the categorical \textit{activity\_type}, which classifies each logged action into one of six mutually exclusive categories:

\begin{itemize}
    \item \textbf{Module (assignment):} Access to a course module or homework set.
    \item \textbf{Reading (reading):} Interaction with instructional content such as textbook pages.
    \item \textbf{Quiz (assessment):} Initiation of a practice or evaluation component.
    \item \textbf{Problem (assessment-questions):} Submission or interaction with individual assessment items. Because each quiz contains multiple questions, this category represents the majority of recorded events.
    \item \textbf{Extra (ancillary):} Access to supplemental instructional materials, such as videos or external references.
    \item \textbf{Hint (promptly:preset):} Automated system feedback events triggered during assessment interactions.
\end{itemize}

The empirical distribution of activity types is highly imbalanced. Problem events account for more than 70\% of all records, while Hint events occur in fewer than 0.2\%. This extreme class imbalance poses challenges for generative modeling, particularly for methods that rely on parametric density assumptions or adversarial training objectives.

The stated requirement of the associated data science workshop was to preserve categorical frequency proportions exactly, including low-frequency tail events, while ensuring that no individual student record could be reconstructed. The objective was structural fidelity of event distributions rather than predictive modeling performance.

Using the training split of the dataset, we apply the Non-Parametric Gaussian Copula (NPGC) synthesizer to generate 10{,}332 synthetic interaction records. Because NPGC models marginal distributions empirically, it preserves the discrete categorical distribution of \textit{activity\_type} without collapsing rare classes.

The resulting synthetic dataset contains 7{,}227 Problem events, 552 Reading events, 27 Extra events, and 16 Hint events, with proportional agreement relative to the original distribution. These counts demonstrate preservation of both dominant and low-frequency categories under augmentation.

The synthetic dataset is released for workshop and research use under privacy constraints. Since NPGC does not reproduce individual-level sequences and operates on an aggregated tabular structure, the released data maintain categorical frequency structure while mitigating identity disclosure risk.

\section{Baseline Hyperparameters}
\label{app:hyper_competitors}

All baseline models were implemented using the SDV single-table synthesizers. Unless otherwise specified, default SDV hyperparameters were used without additional tuning.

Parametric Gaussian Copula (PGC) models each numeric column using a specified parametric family and couples variables through a Gaussian copula correlation structure. The settings below indicate the default numeric marginal family and two post-processing constraints: whether generated values are clipped to the observed range and whether numeric outputs are rounded to match discrete-valued columns.

\begin{table}[hbt]
\centering
\small
\caption{Parametric Gaussian Copula (PGC) Hyperparameters}
\label{tab:pgc_hyper}
\begin{tabular}{lc}
\toprule
\textbf{Parameter} & \textbf{Value} \\ \midrule
Default numerical distribution & beta \\
Enforce min/max values & True \\
Enforce rounding & True \\ \bottomrule
\end{tabular}
\end{table}

Copula-GAN applies a copula-based transformation and then trains a GAN to model dependencies in the transformed space. The hyperparameters below define the training budget (epochs, batch size), representation size (embedding dimension), network capacity (hidden layer sizes), optimizer step sizes (learning rates), discriminator update policy (ste\-ps), and the PAC setting (number of samples packed per discriminator input). The last two options enforce range clipping and rounding during post-processing to better match the original data domain.

\begin{table}[hbt]
\centering
\small
\caption{Copula-GAN Hyperparameters }
\label{tab:copulagan_hyper}
\begin{tabular}{lc}
\toprule
\textbf{Parameter} & \textbf{Value} \\ \midrule
Epochs & 300 \\
Batch size & 500 \\
Embedding dimension & 128 \\
Generator hidden dimensions & (256, 256) \\
Discriminator hidden dimensions & (256, 256) \\
Generator learning rate & $2\times10^{-4}$ \\
Discriminator learning rate & $2\times10^{-4}$ \\
Discriminator steps & 1 \\
PAC & 10 \\
Enforce min/max values & True \\
Enforce rounding & True \\ \bottomrule
\end{tabular}
\end{table}

CTGAN is a conditional GAN designed for mixed continuous and categorical tabular data. The hyperparameters below specify the training budget (epochs, batch size), embedding size for conditional vectors and categorical representations, generator/discriminator capacity (hidden dimensions), optimizer step sizes (learning rates), discriminator update policy (steps per generator update), and PAC packing. As above, min/max enforcement clips generated values to the observed range and rounding enforces discrete-valued outputs where applicable.

\begin{table}[hbt]
\centering
\small
\caption{CTGAN Hyperparameters }
\label{tab:ctgan_hyper}
\begin{tabular}{lc}
\toprule
\textbf{Parameter} & \textbf{Value} \\ \midrule
Epochs & 300 \\
Batch size & 500 \\
Embedding dimension & 128 \\
Generator hidden dimensions & (256, 256) \\
Discriminator hidden dimensions & (256, 256) \\
Generator learning rate & $2\times10^{-4}$ \\
Discriminator learning rate & $2\times10^{-4}$ \\
Discriminator steps & 1 \\
PAC & 10 \\
Enforce min/max values & True \\
Enforce rounding & True \\ \bottomrule
\end{tabular}
\end{table}

TVAE is a variational autoencoder for tabular data. The hyperparameters below specify the training budget (epochs), latent/embedding size, encoder (compression) and decoder (decompression) layer widths, the strength of L2 regularization, and the loss weighting factor used in SDV’s implementation. Min/max enforcement clips outputs to the observed range, and rounding enforces discrete-valued outputs where applicable.

\begin{table}[hbt]
\centering
\small
\caption{TVAE Hyperparameters }
\label{tab:tvae_hyper}
\begin{tabular}{lc}
\toprule
\textbf{Parameter} & \textbf{Value} \\ \midrule
Epochs & 300 \\
Embedding dimension & 128 \\
Compression layers & (128, 128) \\
Decompression layers & (128, 128) \\
L2 regularization scale & $1\times10^{-5}$ \\
Loss factor & 2 \\
Enforce min/max values & True \\
Enforce rounding & True \\ \bottomrule
\end{tabular}
\end{table}

NPGC is a nonparametric Gaussian copula synthesizer. The entries below summarize the privacy level used in experiments (the DP budget $\epsilon$) and the modeling choices: no neural optimization, empirical (nonparametric) marginal anchoring, and dependence captured through a Gaussian copula correlation matrix.

\begin{table}[hbt]
\centering
\small
\caption{NPGC Hyperparameters}
\label{tab:npgc_hyper}
\begin{tabular}{lc}
\toprule
\textbf{Parameter} & \textbf{Value} \\ \midrule
Differential privacy parameter ($\epsilon$) & 1.0 \\
Neural optimization & None \\
Marginal modeling & Empirical\\
Dependency modeling & Gaussian copula relation \\ \bottomrule
\end{tabular}
\end{table}

\subsection{Computational Complexity}
\label{complexity}

Let $n$ denote the number of rows in the training data and $p$ the number of columns. For sampling, let $m$ denote the number of generated rows. The computational cost of the non-parametric Gaussian copula synthesizer can be separated into the fitting stage and the sampling stage.

\paragraph{Fitting complexity}
The algorithm learns the marginal distributions for each column and then estimates the latent Gaussian correlation matrix. For each of the $p$ columns, the marginal-learning step requires sorting or equivalent ranking/uniqueness operations on up to $n$ observations, yielding a cost of order $O(n \log n)$ per column. Across all columns, this contributes
\[
O(pn\log n).
\]
After transforming the data into the latent Gaussian space, the empirical correlation matrix is computed from a matrix of size $n \times p$, which costs
\[
O(np^2).
\]
When differential privacy is enabled, the noisy correlation matrix is projected onto the cone of valid correlation matrices through an eigen-decomposition, which costs
\[
O(p^3).
\]
Therefore, the overall complexity of \texttt{fit()} is
\[
O\!\left(pn\log n + np^2 + p^3\right)
\]
when privacy is enabled, and
\[
O\!\left(pn\log n + np^2\right)
\]
when the correlation-repair step is skipped.

\paragraph{Sampling complexity}
The algorithm first generates an independent Gaussian matrix of size $m \times p$, which costs
\[
O(mp).
\]
It then applies the learned correlation structure using a Cholesky factorization of the $p \times p$ correlation matrix. Computing the factorization costs
\[
O(p^3),
\]
and multiplying the sampled Gaussian matrix by the Choles\-ky factor costs
\[
O(mp^2).
\]
Finally, each column is transformed back through its inverse empirical CDF. For continuous and categorical variables, this step is typically linear in the number of generated samples, contributing approximately
\[
O(mp)
\]
overall. Hence, the typical complexity of \texttt{sample()} is
\[
O\!\left(mp^2 + p^3\right).
\]

\paragraph{Integer-valued columns}
There is, however, an important exception for integer-valued variables when minimum and maximum values are enforced. In that case, the inverse transformation compares each generated value against all unique observed integer levels. If a column has $u_j$ unique integer values, then this step costs
\[
O(mu_j).
\]
In the worst case, $u_j = O(n)$, so a single integer-valued column may require
\[
O(mn).
\]
If $m=n$, this becomes
\[
O(n^2).
\]
Thus, while sampling is usually cheaper than fitting, high-cardinality integer columns can make the sampling stage substantially more expensive in the worst case.

\paragraph{Summary}
Using asymptotic notation, the main computational costs are
\[
\texttt{fit()} = O\!\left(pn\log n + np^2 + p^3\right),
\]
and, in the typical case,
\[
\texttt{sample()} = O\!\left(mp^2 + p^3\right).
\]
With high-cardinality integer columns, the sampling complexity may increase to
\[
O\!\left(mp^2 + p^3 + mn\right).
\]

\paragraph{Memory complexity}
The memory requirements can also be separated into the fitting and sampling stages. During fitting, the algorithm stores the latent Gaussian matrix $Z \in \mathbb{R}^{n \times p}$, which requires
\[
O(np)
\]
memory. It also stores the correlation matrix, which requires
\[
O(p^2),
\]
and the learned marginal information. For numeric variables, the fitted model stores the sorted observed values for each column; across all columns, this requires
\[
O(np)
\]
memory in the worst case. Categorical counts and metadata contribute at most lower-order terms relative to this. Therefore, the overall memory complexity of \texttt{fit()} is
\[
O(np + p^2).
\]

\paragraph{Sampling memory}
During sampling, the algorithm stores the independently generated Gaussian samples, the correlated Gaussian samples, and the final synthetic table, each of size $m \times p$. Although these objects may coexist temporarily, their combined memory usage remains
\[
O(mp)
\]
up to constant factors. The learned correlation matrix again requires
\[
O(p^2)
\]
memory. Hence, the typical memory complexity of \texttt{sample()} is
\[
O(mp + p^2).
\]

\paragraph{Integer-valued columns}
As with runtime complexity, in\-teger-valued columns can increase memory usage. When minimum and maximum values are enforced, the inverse mapping step forms a temporary pairwise difference matrix between the generated values and all unique observed integer levels. If column $j$ has $u_j$ unique integer values, this temporary array requires
\[
O(mu_j)
\]
memory. In the worst case, $u_j = O(n)$, so one such column may require
\[
O(mn)
\]
additional memory. Therefore, the worst-case memory complexity of \texttt{sample()} becomes
\[
O(mp + p^2 + mn).
\]

\paragraph{Summary}
The main memory costs are
\[
\texttt{fit()} = O(np + p^2),
\]
and, in the typical case,
\[
\texttt{sample()} = O(mp + p^2).
\]
With high-cardinality integer-valued columns, the sampling memory complexity may increase to
\[
O(mp + p^2 + mn).
\]

\balancecolumns
\end{document}